\documentclass{article}
\usepackage{arxiv}
\usepackage[utf8]{inputenc} 
\usepackage[T1]{fontenc}
\usepackage{url}            
\usepackage{booktabs}       
\usepackage{amsfonts}       
\usepackage{nicefrac}       
\usepackage{microtype}      
\usepackage{lipsum}         
\usepackage{natbib}
\usepackage{doi}
\usepackage{graphicx}
\usepackage{comment}
\usepackage{multirow}
\usepackage{tabularx,colortbl} 
\usepackage[nointegrals]{wasysym}
\usepackage{float}
\usepackage{rotating}
\usepackage{pdflscape}
\usepackage{amsmath}
\usepackage{hyperref}       
\usepackage{cleveref}       

\usepackage{parskip}

\usepackage{rotating,tabularx}
\usepackage{comment}

\title{U-CFR: Uncertainty-Guided Cascade Forward Refinement for Interactive Segmentation}		

\author{
	\href{https://orcid.org/0009-0002-0414-8371}{\includegraphics[scale=0.06]{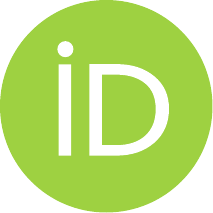}\hspace{1mm}Elijah Danquah Darko}
        \\
		Department of Computer Science\\
		University of Idaho\\
		\texttt{edanquahdarko@uidaho.edu} \\
		\And
		\href{https://orcid.org/0000-0001-6098-4441}{\includegraphics[scale=0.06]{orcid.pdf}\hspace{1mm}Min Xian} \\
		Department of Computer Science\\
		University of Idaho\\
		\texttt{mxian@uidaho.edu} \\
		\And
		\href{https://orcid.org/0000-0001-5572-4303}{\includegraphics[scale=0.06]{orcid.pdf}\hspace{1mm}Terence Soule} \\
		Department of Computer Science\\
		University of Idaho\\
		\texttt{tsoule@uidaho.edu} \\
		\And
		\href{https://orcid.org/0000-0001-8330-7638}{\includegraphics[scale=0.06]{orcid.pdf}\hspace{1mm}Tiankai Yao} \\
		Idaho National Laboratory\\
		\texttt{Tiankai.Yao@inl.gov} \\
		\And
		\href{https://orcid.org/0000-0001-7997-6134}{\includegraphics[scale=0.06]{orcid.pdf}\hspace{1mm}Matthew William Anderson} \\
		Idaho National Laboratory\\
		\texttt{Matthew.Anderson2@inl.gov} \\
	}

\begin{document}
	\maketitle              
	\begin{abstract}
		Interactive image segmentation is critical for efficient image annotation; however, existing methods often require many corrective clicks or rely on passive refinement schemes that converge slowly. We propose Uncertainty-Guided Cascade Forward Refinement (U-CFR), a novel inference-time framework that enables models to autonomously self-correct after each user interaction. U-CFR introduces a boundary-aware uncertainty score that fuses segmentation uncertainty, contour gradients, and explicit edge predictions to guide the placement of internal pseudo-clicks. These self-generated clicks target the most ambiguous boundary regions, providing strong corrective signals without additional manual input. To support this process, we design a dual-head network with a shared encoder-decoder backbone: a segmentation head ensures region consistency, while an edge head sharpens boundary alignment. In inference, U-CFR launches a cascade of refinement steps, where each stage leverages the uncertainty-driven pseudo-clicks to refine the mask progressively. Experiments on standard benchmark datasets demonstrate that the proposed U-CFR improves click efficiency, initial mask quality, and boundary accuracy. It reduces the required clicks by over 10\% on challenging datasets like Berkeley and offers a more intelligent and efficient interactive annotation.
		\keywords{Deep Learning \and Interactive Image Segmentation \and Interactive Refinement.}
	\end{abstract}
	
\section{Introduction}
	\label{sec:intro}
	Interactive image segmentation stands as a critical task in computer vision, forming the foundation for a multitude of applications, ranging from efficient data annotation for large-scale datasets to precise medical image analysis \cite{liu2023simpleclick}. The fundamental objective is to generate segmentation masks at the pixel-level that accurately delineate an object of interest, guided by minimal and intuitive user interactions, typically in the form of clicks \cite{xian2016eiseg}, scribbles \cite{xian2016neutro,liu2023simpleclick,sun2024cfr}, bounding boxes \cite{xu2017deep}, or a combination \cite{zhang2020interactive} of these. In recent years, the field has witnessed remarkable progress, driven by the advent of powerful deep learning architectures. This evolution has culminated in two primary classes of models: generalist foundation models, such as Segment Anything Model (SAM) \cite{kirillov2023segment} and SEEM \cite{zou2023segment}, which have demonstrated impressive zero-shot generalization capabilities across a vast range of domains and specialist models \cite{liu2023simpleclick,chen2022focalclick,sun2024cfr,sofiiuk2021reviving}, explicitly optimized for the segmentation task, continue to provide state-of-the-art performance in resource-constrained environments. This paper focuses on the latter paradigm.
	Despite these significant advancements, a persistent and formidable challenge remains. (1) They cannot accurately segment objects characterized by complex topologies. These objects are often present with fine-grained details, intricate and non-convex boundaries, or thin structures that are only a few pixels wide \cite{liew2021deep}. (2) Small objects, which occupy a minimal pixel area within an image, are similarly difficult to segment accurately. Standard deep learning architectures, particularly those employing successive downsampling operations, often lose high-frequency spatial information necessary to resolve such fine details, leading to the features of small objects being diluted or lost entirely within the broader context scene \cite{sang2022small}. These failure modes are not merely academic cases; they represent a significant bottleneck in practical applications where precision is paramount \cite{liew2021deep}. This paper addresses these specific shortcomings in current state-of-the-art methods.

	\noindent State-of-the-art methods like SimpleClick \cite{liu2023simpleclick}  have demonstrated the power of plain Vision Transformer (ViT) \cite{dosovitskiy2020image}, but their lightweight decoders can struggle with fine-grained details. Other approaches, such as CFR-ICL \cite{sun2024cfr}, introduce an inference-time refinement loop. These are often unguided, applying refinement effort uniformly across the image rather than focusing on critical error regions or boundaries. This inefficiency highlights the need for more intelligent refinement strategies that autonomously identify and correct errors in a targeted manner.

	\noindent To address the limitations, we propose Uncertainty-Guided Cascade Forward Refinement (U-CFR)\footnote{Code available at \url{https://github.com/elidandar/UCFR-Interactive-Segmentation}}, a novel framework that enables a model to self-correct after each user interaction. Although related to PseudoClick generation, our work differs from PseudoClick \cite{liu2022interactive} and CFR-ICL \cite{sun2024cfr} in two key ways. \textbf{First}, our method is built on a dual-head architecture that incorporates an auxiliary edge-detection head. This provides a dedicated stream of high-fidelity boundary features, which is critical to the refinement goal. \textbf{Second}, our uncertainty map fuses predictive uncertainty with explicit contour gradients, allowing our pseudo-clicks to target ambiguous boundaries, rather than just ambiguous regions. By integrating these components, the proposed framework transforms the refinement process from a passive, uniform operation into an active, intelligent self-correction loop. We demonstrate that this approach achieves competitive results and significantly reduces the number of user clicks for high-quality segmentation on benchmarks. 
	
\section{Related Works}
	\label{sec:relatedworks}
	\textbf{Vision Transformers in Interactive Image Segmentation.} The architectural backbone is a cornerstone of any interactive segmentation model, responsible for extracting rich and informative features from the input image and user prompts. 
	Early deep learning-based methods predominantly relied on Convolutional Neural Networks (CNNs), such as ResNet \cite{he2016deep} and HRNet \cite{wang2020deep}, which use hierarchical structures to build receptive fields and capture multi-scale information. The advent of the Vision Transformer (ViT) \cite{dosovitskiy2020image} marked a significant turning point. 
	These models demonstrated strong performance by combining the global context modeling of self-attention with the multi-scale feature representations of CNNs. The introduction of SimpleClick \cite{liu2023simpleclick} pioneered a more radical approach, demonstrating the effectiveness of a plain, non-hierarchical ViT architecture. A plain ViT can capture global spatial relationships from the first self-attention block, in contrast to the progressively expanding receptive fields of CNNs. Furthermore, this simpler and more uniform architecture is highly compatible with advanced self-supervised pretraining methods such as masked autoencoders (MAE) \cite{he2022masked}, which have been shown to learn powerful and generalizable visual representations. The success of SimpleClick \cite{liu2023simpleclick} validated this approach. 
	Current state-of-the-art continues to be advanced by Transformer-based models, with recent works, such as CFR-ICL \cite{sun2024cfr}, Order-Aware Interactive Segmentation (OIS) \cite{wang2024order}, and MST \cite{xu2025mst}.

	\noindent\textbf{Boundary-Aware and Multi-Task Segmentation.} The challenge of achieving precise segmentation along object boundaries is a long-standing problem. A well-established and effective strategy to address this is to explicitly incorporate boundary or edge information into the model's learning process. This is typically framed as a multi-task learning problem, where a shared encoder backbone feeds into two or more distinct decoder heads: one for the primary task of semantic segmentation and an auxiliary head for the secondary task of edge detection \cite{khattar2021cross}.
	The underlying principle of this approach is that the explicit supervision of the edge detection task acts as a regularizer on the shared feature representations. It forces the encoder to learn features that are not only semantically informative but also spatially precise and respectful of object boundaries. This dual objective helps to mitigate the tendency of standard segmentation models to produce smooth predictions that blur fine details. The success of this approach is not limited to CNNs; recent works such as the Boundary-Aware Transformer (BAT) \cite{wang2021boundary} has demonstrated its successful application to Transformer-based architectures, proving the principle's relevance to our proposed framework.
	A critical aspect of implementing such a dual-task system is designing the loss function for the edge-prediction branch. Edge maps are inherently sparse, leading to a severe class imbalance between the small number of edge pixels and the vast majority of non-edge pixels. Common practice involves using a class-balanced Binary Cross-Entropy (BCE) loss, the Dice loss \cite{diceloss}, or their combination \cite{deng2018learning}.
    
	\noindent\textbf{Refinement Strategies in Segmentation.} Even with powerful backbones and optimized training schemes, the initial output of a segmentation model is often a coarse approximation of the true object mask. Consequently, a variety of refinement strategies have been developed to post-process these initial predictions and improve their quality. These strategies can be broadly categorized into model-agnostic and model-integrated approaches. Model-agnostic methods operate as a separate post-processing step, taking the coarse mask and the original image as input. A notable example is SegRefiner \cite{wang2023segrefiner}, which frames the refinement task as an iterative data generation process, using a denoising diffusion model to progressively correct errors in the mask. While powerful, these methods add a distinct and often computationally intensive stage to the inference pipeline.
	Model-integrated approaches, by contrast, build the refinement mechanism directly into the primary model's architecture or inference loop. Cascade-Forward Refinement (CFR), introduced in CFR-ICL \cite{sun2024cfr}  by Sun et. al, is a prime example of an efficient inference-time iterative method that does not require additional trainable modules. A particularly innovative direction within model-integrated refinement is click imitation or "pseudo-click" generation. Pioneered by Lui et. al and presented in PseudoClick \cite{liu2022interactive}, this approach enables the segmentation network itself during training to propose the next corrective click by identifying error regions in its own output. These automatically generated clicks serve as a proxy for human interaction, allowing the model to correct its predictions during training.
	
\section{Method}
	\subsection{Architectural Overview}
\begin{figure*}[t]
		\centering
		\includegraphics[width=0.9\textwidth, height=7cm]{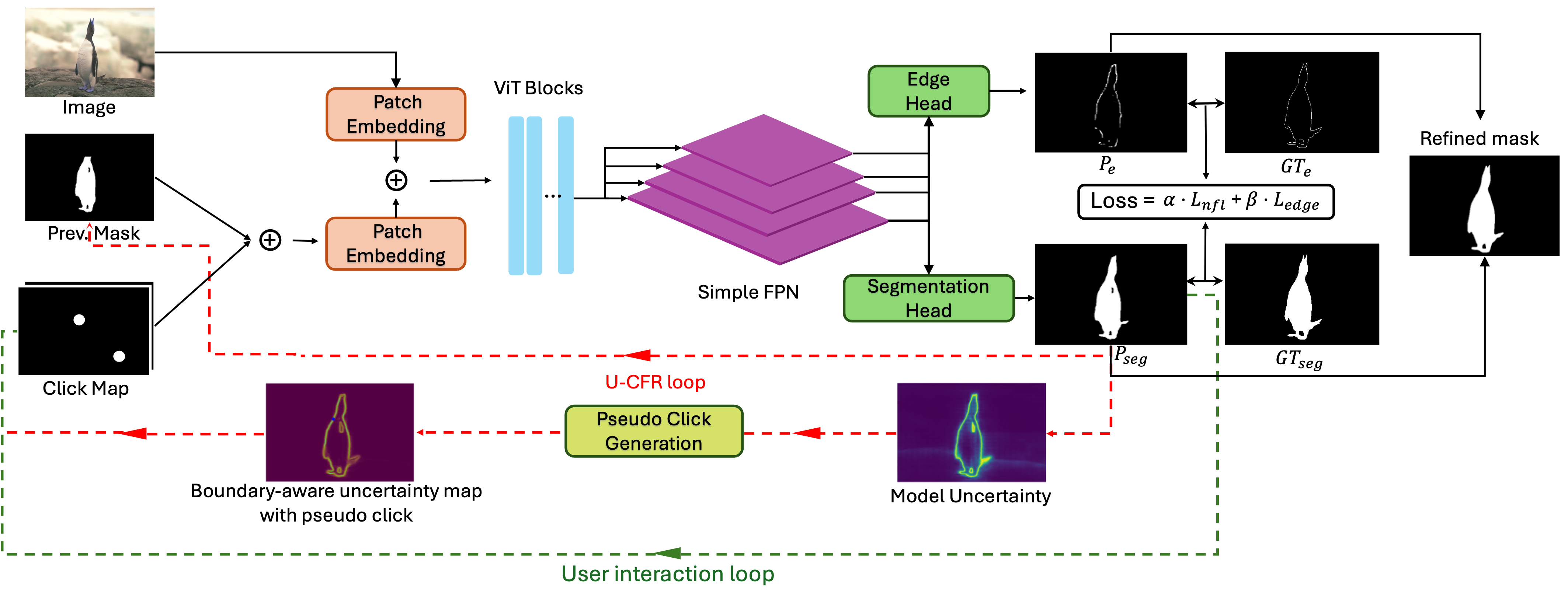}
		\caption{
			Overview of the proposed Uncertainty-Guided Cascade Forward Refinement (U-CFR) framework. User-provided clicks (middle left) are embedded into click maps and fused with the input image before entering a shared encoder-decoder backbone.
		}
		\label{fig:model}
		
	\end{figure*}

	We propose a deep multitask learning framework for interactive image segmentation, which consists of a region segmentation branch and an edge detection branch. The edge detection branch is introduced to address the challenges of inaccurate object boundaries in interactive image segmentation. In addition, a novel refinement strategy is proposed to incorporate an uncertainty-guided pseudo-clicks generation into the CFR \cite{sun2024cfr} loop. It generates meaningful clicks in highly uncertain image regions, leading to more accurate segmentation without requiring additional real user interactions. The architecture of the proposed method is shown in \Cref{fig:model}.
	
	\noindent\textbf{Shared Backbone and Encoding Clicks.} The backbone of the proposed architecture mainly consists of a shared plain ViT backbone (pretrained with MAE) \cite{he2022masked}. To handle objects at various scales, the features from the final ViT block are aggregated by a simple Feature Pyramid Network (FPN) \cite{lin2017feature}, which generates multi-scale features for the decoder heads. The backbone encodes the input image ($
	H\times W\times3$) and the user clicks ($
	H\times W\times3$). User clicks (positive and negative clicks) are encoded into a two-channel disk map and concatenated with the segmentation mask from the previous iteration. This three-channel map is processed by a symmetric patch embedding layer, and its output is added element-wise to the embeddings of the image patches. The resulting sequence of tokens is fed through the ViT blocks to the FPN to produce a multi-scale, high-dimensional feature map, $F_{backbone}$, which encapsulates a rich, global understanding of the image.

	\noindent\textbf{Edge Detection Head}. A new, lightweight decoder branch is added in parallel to the main segmentation path. It takes the same backbone feature map, $F_{backbone}$, as input and is specifically tasked with producing a probability map of the object's boundaries, denoted as $P_{e}$.
	
	\noindent\textbf{Uncertainty Guided CFR}. This is a novel, inference-time iterative process to feed back the segmentation results and generate pseudo-clicks. It takes the initial predictions from the segmentation head ($P_{seg}$) 
	to calculate a guidance map. This map then dynamically modulates the features within an iterative refinement loop to produce the final segmentation mask.
	
	\subsection{Edge Detection Head (EDH) as Auxiliary Supervision}
	The EDH is designed to inculcate the shared ViT backbone with an explicit awareness of high-frequency boundary details, acting as an auxiliary supervisory signal during the training process. It is a lightweight decoder that mirrors the structural simplicity of the segmentation head. This structure efficiently transforms the coarse, high-dimensional feature map into a full-resolution, single-channel probability map, $P_{edge}$, where each pixel value represents the likelihood of it being part of an object boundary. This takes the final feature map from $F_{backbone}$ as input, similar to the segmentation head.
	
	\noindent\textbf{Ground Truth Generation}. A key practical advantage of this approach is that it does not require expensive, manually annotated edge labels. Instead, the ground-truth edge maps, $GT_{e}$, are generated on-the-fly during training directly from the available ground truth masks of object regions, $GT_{seg}$. This is achieved by applying a standard differential image processing operator, such as a Laplacian or Sobel filter \cite{kanopoulos1988design}, to the binary segmentation mask. The result is a single-pixel-wide contour representing the object's boundary, which serves as the supervision target for the EDH.

	
	\noindent\textbf{Loss Function}
	The training of the EDH is supervised by a dedicated edge loss, $L_{edge}$. Due to the severe class imbalance inherent in edge maps (where edge pixels represent a small fraction of all pixels), we employ the Dice loss~\cite{sudre2017generalised}, which is robust to such imbalance by construction. While topologically-aware variants such as clDice~\cite{shit2021cldice} could better capture the connectivity of thin boundary structures, standard Dice loss is sufficient here since the EDH serves as a training-time regulariser rather than a standalone boundary detector. Its role is to encourage boundary-sensitive feature representations in the shared backbone, not to produce geometrically precise edge maps. The total training objective is a weighted sum of the primary segmentation loss and this auxiliary edge loss.

	\begin{equation}\label{eq:L_total}
		\begin{aligned}
			L_{total}     & =\alpha \cdot L_{nfl} + \beta \cdot L_{edge}
		\end{aligned}
	\end{equation}
	
	\begin{equation}\label{eq:L_edge}
		\begin{aligned}
			L_{edge} = 1 - \frac{2 \cdot \sum(GT_{e} \cdot P_{e})}{\sum GT_{e} + \sum P_{e}}
		\end{aligned}
	\end{equation}
	
	\begin{equation}\label{eq:L_nfl}
		\begin{aligned}
			L_{nfl} &= \sum\nolimits\left(- \frac{(1 - P_{seg}(i,j)^\gamma\log{P_{seg}(i,j)}}{|P_{seg}|}\right)
		\end{aligned}
	\end{equation}

	The loss term $L_{nfl}$ is the normalized focal loss \cite{sofiiuk2019adaptis} for the segmentation head, designed for the imbalanced nature of interactive segmentation. $P_{seg}(i,j)$ is the probability prediction at point $(i,j)$, $\gamma>0$ is the tunable parameter, and $ \alpha$ and $\beta$ balance the contribution of the two tasks during optimization, and are selected by experiments.

	\subsection{Uncertainty Guided Cascade-Forward Refinement (U-CFR)} 
	The CFR-based approach \cite{sun2024cfr} is one of the most successful inference strategies employed in interactive image segmentation, improving the quality of the segmentation during inference in a coarse-to-fine manner. It introduced two inference loops to generate segmentation masks using incremental user interactions, and forward the segmentation model multiple times with the same input image and user clicks. CFR generates promising overall results, but primarily depends on user clicks and still needs intensive user interactions for challenging images.
	
	We propose U-CFR, an improved inference-time strategy that introduces uncertainty-guided pseudo-clicks into the CFR framework. The pseudo-click generation is proposed to produce more informative clicks that help the segmentation backbone distinguish highly uncertain image regions. We produce the pseudo-clicks by using a boundary-aware uncertainty map calculated using the predictive uncertainty and contour gradients.
	
	\noindent \textbf{Boundary-Aware Uncertainty}. The core of our guidance strategy is a map that identifies pixels that are simultaneously uncertain and located on the boundary of the object. This map, $U_{bd}$, is computed by fusing two signals:
	\begin{itemize}
		\item \textbf{Predictive uncertainty.} A standard uncertainty map, $U_{pred}$, is computed from the segmentation probability map $P_{seg}$. It is defined by 
		\begin{equation}\label{eq:uncertainty}
			\begin{aligned}
				U_{pred} &= 1 - | 2  \cdot  P_{seg}- 1  |, \\
			\end{aligned}
		\end{equation}
		
		which assigns high values to pixels where the predicted probability is close to 0.5.
		
		\item \textbf{Contour gradients.} The normalized gradient magnitude of the segmentation probability map,
		\begin{equation}\label{eq:gradient}
			\begin{aligned}
				G_{seg} &=  | \nabla P_{seg} |, \\
			\end{aligned}
		\end{equation}
		is computed using a Sobel filter \cite{kanopoulos1988design}. This highlights regions where the segmentation prediction changes sharply, effectively outlining the predicted object's contour.
		
		\noindent These signals from \Cref{eq:gradient,eq:uncertainty} are fused to create the final boundary-aware uncertainty map  
		
		\begin{equation}\label{eq:uncertain_boundary}
			\begin{aligned}
				U_{bd} &=  U_{pred} \cdot G_{seg}. \\
			\end{aligned}
		\end{equation}
		The resulting map has high values only in regions that are both on a predicted boundary and where the model's segmentation confidence is low.
	\end{itemize}
	
	\noindent \textbf{Uncertainty-Guided Pseudo-Click Generation.} U-CFR operates in an iterative loop for a fixed number of steps after each user interaction.  In each step, the coordinates $(i^\star,j^\star)$ of the pixel with the maximum value in the boundary-aware uncertainty map ($argmax$ $U_{bd}$) is identified as the candidate pseudo-click location. 
	To assign a label to a potential pseudo-click, we adopt a selective confidence rule based on the model's output probability, $P_{seg}$. This rule uses two thresholds, a lower bound $\tau^-$ and an upper bound $\tau^+$ to categorize the model's confidence as shown in \Cref{eq:rule}
	\begin{equation}\label{eq:rule}
		label(i^\star,j^\star) =  \begin{cases}
			\text{skip}, \quad  \tau^- \leq P_{seg}(i^\star,j^\star) \leq \tau^+  \\
			\text{positive}, \quad  P_{seg}(i^\star,j^\star) > \tau^+  \\
			\text{negative}, \quad  P_{seg}(i^\star,j^\star) < \tau^-  \\
			
		\end{cases}
	\end{equation}
	This rule deliberately skips placing a pseudo-click in the ambiguous interval between $\tau^-$ and $\tau^+$, where the model's prediction is least certain and essentially noise.  Instead, it only generates a positive or negative pseudo-click when the model has high confidence that its prediction is correct $(>\tau^+)$ or incorrect $(<\tau^-)$. From our experiments we set $\tau^-$ and $\tau^+$ to $0.49$ and $0.51$ respectively. This strategy mirrors human annotator behavior: corrections are typically made in regions that are clearly wrong, not in regions of complete ambiguity. This ensures that each generated pseudo-click contributes effectively to the refinement process. Once a "positive" or "negative" label is assigned, the new pseudo-click is added to the set of existing clicks. This updated set is then fed back into the model to generate a refined segmentation mask. This iterative process allows the model to autonomously correct its own most significant errors, leading to higher-quality final masks with fewer manual interactions.
	
\section{Experiments}
	To evaluate the efficacy of the proposed framework, a comprehensive experimental study is designed. This study includes analyses to isolate the contributions of each component, followed by a direct quantitative and qualitative comparison against a suite of state-of-the-art interactive segmentation models.
	\subsection{Setup}
	\noindent\textbf{Datasets}: To ensure fair and direct comparability with existing foundational works, we evaluate our framework on standard datasets of public benchmarks widely used in the interactive segmentation literature. These include Semantic Boundaries Dataset (SBD) \cite{hariharan2011semantic}, COCO\_MVal \cite{lin2014microsoft}, Pascal VOC \cite{everingham2010pascal}, DAVIS \cite{perazzi2016benchmark}, Berkeley \cite{martin2001database}, GrabCut \cite{rother2004grabcut}, Brain Tumor Segmentation challenge (BraTS) \cite{baid2021rsna}, ssTEM \cite{gerhard2013segmented} and OAIZIB \cite{ambellan2019automated}. For training, we selected SBD \cite{hariharan2011semantic}, 
	which consists of 8,498 training images and 2,820 test images.\\ 
    
	\noindent \textbf{Evaluation Metrics}: The primary metric for assessing performance will be the Number of Clicks (NoC) required to achieve a target Intersection over Union (IoU) of 85\% and 90\%, denoted as NoC@85 and NoC@90, respectively. This metric directly measures the user effort required to obtain a high-quality mask and is the standard for comparing interactive segmentation methods \cite{sun2024cfr,liu2023simpleclick,lin2022focuscut,liu2022interactive}. To analyze convergence behavior, the mean IoU after a fixed number of clicks (mIoU@k) will also be reported. To specifically quantify the improvement in boundary segmentation, we utilize 
	Normalized Surface Distance Score (NSDS) \cite{kavur2021chaos}. 
	NSDS is sensitive to the accuracy of the segmentation boundaries as it measures the distance between the surfaces of the predicted segmentation and the ground truth. A higher score indicates a better boundary alignment.

	\noindent \textbf{Implementation Details}: We implement the model using Python and PyTorch~\cite{paszke2019pytorch}. The pretrained ViT-Base (ViT-B) model is adopted, which uses a Plain Vision Transformer \cite{dosovitskiy2020image} as its backbone. The model is fine-tuned for 45 epochs on top of the provided SimpleClick \cite{liu2023simpleclick} weights using our proposed methods. For optimization, we employ the Adam optimizer with parameters $\beta_1=0.9, \beta_2=0.999$, a learning rate of $5\times10^{-6}$, and a batch size of $140$. Our training objective combines two loss functions: for the segmentation task, we use Normalized Focal Loss (NFL) \cite{sofiiuk2019adaptis} with $\alpha=0.5$ and $\gamma=2$, while the edge prediction task is supervised with Dice Loss \cite{diceloss}. To emphasize the importance of learning precise boundary features from the auxiliary task, we set the weights at $\alpha=0.2$ and $\beta=2$. This higher weight on edge loss encourages the model to develop robust edge-aware features, which in turn benefit the final segmentation quality. The input images are resized to $448\times448$. We apply a range of data augmentations, including random resizing, flipping, rotation, cropping, and adjustments to brightness and contrast. User clicks are encoded into disk maps with a radius of 5 pixels. All experiments are conducted on an NVIDIA  Quadro RTX A6000. 
	
	\subsection{The Effectiveness of Edge Detection Head}
	To isolate the contribution of the auxiliary edge detection head (EDH), we compare the dual-head architecture with the SimpleClick baseline \cite{liu2023simpleclick}, which functions as a model trained with only the segmentation head. Both models share the same encoder and segmentation head design to ensure a fair comparison. 
	
	The results, detailed in \Cref{tab:effectiveness_epd}, demonstrate that the inclusion of the EDH provides consistent and meaningful improvements. First, we observe an improvement in click efficiency. For the NoC@90 metric, our Dual Head model reduces the required clicks by 3.9\% on PascalVOC (from 2.81 to 2.70), 4.2\% on COCO\_MVal (from 4.07 to 3.90), and 4.4\% on SBD (from 5.24 to 5.01). This trend continues for the more stringent NoC@95 metric, indicating that the boundary-aware features learned by the EDH help the model converge on a high-quality mask with fewer user interactions.
	
	Beyond click efficiency, the EDH significantly enhances segmentation quality, especially in the first few clicks. On the COCO\_MVal dataset, our model improves the initial mIoU by 8.2\% (from 61.03\% to 66.05\%) and the initial NSDS by 6.9\% (from 0.29 to 0.31). This performance gain is consistent across other datasets. On PascalVOC, our model achieves a higher mIoU after just one click (79.66\% vs. 78.49\%) and shows a  9.7\% improvement in the boundary-specific NSDS metric at the first click (0.34 vs. 0.31).
	
	While the initial mIoU gains on the SBD dataset are modest, the NSDS  after five clicks improves from 0.65 to 0.66 (a 1.5\% relative improvement), highlighting the benefits of the auxiliary edge task for refining segmentation boundaries. In nearly all scenarios, our model matches or exceeds the baseline, with the most significant advantages seen in boundary accuracy (NSDS) and initial segmentation quality (mIoU@1). These results strongly suggest that explicitly supervising the model on edge detection through the EDH provides valuable features that lead to a more efficient and accurate interactive segmentation process.
	
	\begin{table}[t]
		\centering
		\setlength{\tabcolsep}{2.5pt} 
		\begin{tabular}{@{}l cc cc cc@{}}
			\toprule
			\multirow{2.5}{*}{\textbf{Metric}} & \multicolumn{2}{c}{\textbf{PascalVOC}} & \multicolumn{2}{c}{\textbf{COCO\_MVal}} & \multicolumn{2}{c}{\textbf{SBD}} \\
			\cmidrule(lr){2-3} \cmidrule(lr){4-5} \cmidrule(lr){6-7}
			& Base & Ours & Base & Ours & Base & Ours \\
			\midrule
			NoC@90 $\downarrow$ & 2.81 & \textbf{2.70} & 4.07 & \textbf{3.90} & 5.24 & \textbf{5.01} \\
			NoC@95 $\downarrow$ & 3.75 & \textbf{3.62} & 7.89 & \textbf{7.39} & 11.24 & \textbf{10.79} \\
			\midrule
			mIoU@1 $\uparrow$ & 78.49 & \textbf{79.66} & 61.03 & \textbf{66.05} & 74.28 & \textbf{74.90} \\
			mIoU@5 $\uparrow$ & 94.98 & \textbf{95.56} & 90.37 & \textbf{91.15} & 90.49 & \textbf{90.62} \\
			mIoU@10 $\uparrow$ & 97.63 & \textbf{97.99} & 94.82 & \textbf{95.37} & 92.66 & \textbf{93.02} \\
			\midrule
			NSDS@1 $\uparrow$ & 0.31 & \textbf{0.34} & 0.29 & \textbf{0.31} & 0.41 & 0.41 \\
			NSDS@5 $\uparrow$ & 0.38 & \textbf{0.40} & 0.60 & \textbf{0.61} & 0.65 & \textbf{0.66} \\
			NSDS@10 $\uparrow$ & 0.39 & \textbf{0.40} & 0.72 & \textbf{0.74} & 0.72 & \textbf{0.73} \\
			\bottomrule
		\end{tabular}
		\caption{Comparison of the Dual Head model (Ours) against the SimpleClick(without EDH) baseline (Base). Arrows indicate if lower ($\downarrow$) or higher ($\uparrow$) is better. The EDH consistently improves click efficiency (NoC) and enhances initial segmentation quality (mIoU@1) and boundary accuracy (NSDS).}
		\label{tab:effectiveness_epd}
	\end{table}
	
	\subsection{The Effectiveness of U-CFR}
	\begin{table}[!ht]
		\centering
		\footnotesize 
		\setlength{\tabcolsep}{2.5pt} 
		\begin{tabular}{@{}ll c cc cc@{}}
			\toprule
			\multirow{2}{*}{\textbf{Dataset}} & \multirow{2}{*}{\textbf{Method}} & \textbf{NoC} & \multicolumn{2}{c}{\textbf{mIoU}} & \multicolumn{2}{c}{\textbf{NSDS}} \\
			\cmidrule(lr){3-3} \cmidrule(lr){4-5} \cmidrule(lr){6-7}
			& & \textbf{@90} $\downarrow$ & \textbf{@5} $\uparrow$ & \textbf{@10} $\uparrow$ & \textbf{@10} $\uparrow$ & \textbf{@20} $\uparrow$ \\
			\midrule
			\multirow{2}{*}{PascalVOC} & w/ CFR1 & 2.72 & 95.29 & \textbf{97.94} & 0.40 & 0.40 \\
			& \textbf{w/ U-CFR1} & \textbf{2.64} & \textbf{95.38} & 97.80 & 0.40 & 0.40 \\
			\midrule
			\multirow{2}{*}{COCO\_MVal} & w/ CFR1 & 3.88 & 89.39 & \textbf{95.11} & 0.73 & \textbf{0.83} \\
			& \textbf{w/ U-CFR1} & \textbf{3.80} & \textbf{89.93} & 94.92 & 0.73 & 0.82 \\
			\midrule
			\multirow{2}{*}{SBD} & w/ CFR1 & 4.98 & 90.28 & \textbf{92.85} & 0.73 & 0.80 \\
			& \textbf{w/ U-CFR1} & \textbf{4.94} & \textbf{90.36} & 92.72 & 0.73 & \textbf{0.81} \\
			\bottomrule
		\end{tabular}
		\caption{Effectiveness of U-CFR. We compare our full method (w/ U-CFR) against the baseline refinement (w/ CFR). CFR1 denotes a single refinement step. U-CFR consistently improves click efficiency (NoC) and initial segmentation accuracy (mIoU@5).}
		\label{tab:ucfr_effectiveness}
	\end{table}

	To validate the effectiveness of the U-CFR framework, we conducted a series of experiments to demonstrate that U-CFR can reduce the number of user clicks required to achieve high-quality segmentation. We compare the full model equipped with U-CFR, Ours (base+U-CFR1), against the baseline using Cascade Forward Refinement, Ours (base+CFR1), on PascalVOC, COCO\_MVal, and SBD datasets.
	
	The primary finding, detailed in \Cref{tab:ucfr_effectiveness}, shows U-CFR consistently improves interaction efficiency. On PascalVOC, U-CFR reduces the NoC@90 from 2.72 to 2.64, a relative improvement of 2.9\%. This efficiency gain is consistent across datasets, with a 2.1\% reduction in clicks on COCO\_MVal (from 3.88 to 3.80) and a similar trend on SBD. Furthermore, U-CFR demonstrates an advantage in early stages of interaction. The mIoU after 5 clicks (mIoU@5) is consistently higher, showing an improvement of 0.54 percentage points on COCO MVal (89.93\% vs. 89.39\%) and a similar gain on SBD. This indicates that our active, uncertainty-guided pseudo-clicks provide a more effective corrective signal. While the CFR baseline eventually achieves comparable accuracy after 10 clicks, U-CFR yields a high-quality result sooner, directly contributing to a reduction in user annotation effort. 
	
	\begin{sidewaystable}
		\scriptsize
		\centering
		\begin{tabularx}{\linewidth}{l l c c c c c c c c c c c c c c c}
			\toprule
			\multirow{2}{*}{Method} & \multirow{2}{*}{Backbone} & 
			\multicolumn{3}{c}{GrabCut} & \multicolumn{3}{c}{Berkeley} & \multicolumn{3}{c}{SBD} & \multicolumn{3}{c}{DAVIS} & \multicolumn{3}{c}{Pascal VOC} \\ 
			& & NoC85 & NoC90 & NoC95 & NoC85 & NoC90 & NoC95 & NoC85 & NoC90 & NoC95 & NoC85 & NoC90& NoC95 & NoC85 & NoC90 & NoC95 \\
			\midrule
			LD \cite{li2018interactive} & VGG-19
			& 3.20 & 4.79 & - & - & - & - &  7.41 & 10.78 & - & 5.05 & 9.57 & - & - & - & - \\   & & & & & & & & & & & & & & & & \\
			BRS  \cite{jang2019interactive} & DenseNet
			& 2.60 & 3.60 &  - & - & 5.08 & - & 6.59 & 9.78 & - & 5.58 & 8.24  & - & - & - & - \\  & & & & & & & & & & & & & & & &\\
			f-BRS  \cite{sofiiuk2020f} & ResNet-101
			& 2.30 & 2.72&   & - & 4.57 & - & 4.81 & 7.73 & -  & 5.04 & 7.41 & -  & - & - & - \\   & & & & & & & & & & & & & & & & \\
			RITM~\cite{sofiiuk2021reviving}  & HRNet-18
			& 1.76 & 2.04 & - & 1.87 & 3.22 & - & 3.39 & 5.43 & - & 4.94 & 6.71 & - & 2.51 & 3.03 & - \\    & & & & & & & & & & & & & & & & \\
			CDNet~\cite{chen2021conditional}  & ResNet-34
			& 1.86 & 2.18 & - & 1.95 & 3.27 & - & 5.18 & 7.89 & - & 5.00 & 6.89 & - & 3.61 & 4.51 & - \\   & & & & & & & & & & & & & & & & \\
			PseudoClick~\cite{liu2022interactive} & HRNet-18
			& 1.68 & 2.04 & - & 1.85 & 3.23 & - & 3.38 & 5.40 & - & 4.81 & 6.57  & - & 2.34 & 2.74 & - \\   & & & & & & & & & & & & & & & &\\
			FocalClick~\cite{chen2022focalclick} & HRNet-18s 
			& 1.86 & 2.06 & - & - & 3.14 &   & 4.30 & 6.52 & - & 4.92 & 6.48 & - & - & - & - \\  & & & & & & & & & & & & & & & & \\
			FocalClick~\cite{chen2022focalclick} & SegF-B0 
			& 1.66 & 1.90 & - & - & 3.14 & - & 4.34 & 6.51 & - & 5.02 & 7.06 & - & - & - & - \\  & & & & & & & & & & & & & & & & \\
			FocusCut~\cite{lin2022focuscut} & ResNet-50 
			& 1.60 & 1.78 & - & 1.85$^\ast$ & 3.44 & - & 3.62 & 5.66 & - & 5.00 & 6.38 & - & - & - & - \\  & & & & & & & & & & & & & & & & \\
			FocusCut~\cite{lin2022focuscut} & ResNet-101
			& 1.46 & 1.64 & - & 1.81$^\ast$ & 3.01 & - & 3.40 & 5.31 & - & 4.85 & 6.22 & - & - & - & - \\   & & & & & & & & & & & & & & & & \\
			GPCIS \cite{zhou2023interactive} & SegF-B0
			& 1.60 & 1.76 & - & 1.84 & 2.70 & - & 4.16 & 6.28 & - & 4.45 & 6.04 & - & - & - & - \\  & & & & & & & & & & & & & & & & \\
			FCFI \cite{wei2023focused} & ResNet101
			& 1.64 & 1.80 & - & - & 2.84 & - & 3.26 & 5.35 & - & 4.75 & 6.48 & - & - & - & - \\  & & & & & & & & & & & & & & & & \\
			EMC \cite{du2023efficient} & HRNet18
			& 1.74 & 1.84 & - & - & 3.03 & - & 3.38 & 5.51 & - & 5.05 & 6.71 & - & 2.37 & - & - \\  & & & & & & & & & & & & & & & & \\
			FDRN \cite{zeng2023feature} & SegF-B0
			& 1.58 & 1.78 & - & - & 3.08 & - & 4.18 & 6.20 & - & 4.78 & 6.66 & - & - & - & - \\  & & & & & & & & & & & & & & & & \\
			SimpleClick \cite{liu2023simpleclick} & ViT-B 
			& 1.40 & 1.54 & 2.16$^\ast$  & 1.44 & 2.46 & 6.71$^\ast$ & 3.28 & 5.24 & 11.24$^\ast$ & 4.10 & 5.48 & 12.23$^\ast$  & 2.38 & 2.81 & 3.75$^\ast$ \\ & & & & & & & & & & & & & & & & \\
			\rowcolor[gray]{0.9}    
			Ours (Base) & ViT-B 
			& \textbf{1.38} & 1.52 & 2.02 & 1.60 & 2.34 & 6.44 & 3.19 & 5.01 & 10.79 & 4.43 & 5.73 & 11.66 & \textbf{2.27} & 2.70 & 3.62 \\  & & & & & & & & & & & & & & & & \\
			Ours (Base+CFR1\cite{sun2024cfr}) & ViT-B 
			& 1.38 & 1.56 & 1.96 & 1.48 & 2.27 & 6.25 & 3.18 & 4.98 & \textbf{10.68} & 4.49 & 5.71 & \textbf{11.56} & 2.31 & 2.72 & 3.57 \\  & & & & & & & & & & & & & & & & \\
			\rowcolor[gray]{0.9} Ours (Base+U-CFR1) & ViT-B 
			& 1.40 & \textbf{1.50} & \textbf{1.82} & \textbf{1.42} & \textbf{2.19} & \textbf{6.07} & \textbf{3.14} & \textbf{4.94} & 10.71 & \textbf{4.29} & \textbf{5.54} & 11.60 & 2.27 & \textbf{2.64} & \textbf{3.50} \\		
			\bottomrule
		\end{tabularx}
		\caption{\textbf{Comparison with state-of-the-art methods.} We report results on five benchmarks: GrabCut \cite{rother2004grabcut}, Berkeley \cite{martin2001database}, SBD~\cite{hariharan2011semantic}, DAVIS~\cite{perazzi2016benchmark}, and Pascal VOC~\cite{everingham2010pascal}. We report the Number of Clicks (NoC) required to reach 85\%, 90\%, and 95\% IoU. CFR1 denotes one step refinement. All models were trained on SBD dataset. Our method, particularly the Ours (Base+U-CFR1) variant, establishes a new state-of-the-art across most datasets. Bold indicates the best performance. * denotes results reproduced by the released model.} 
	\label{tab:comparision_sota_noc}
\end{sidewaystable}

\begin{table}[!ht]
	\centering
	\footnotesize
	\setlength{\tabcolsep}{1.6pt} 
	\begin{tabular}{l l cc ccc cc}
		\toprule
		\multirow{2}{*}{\textbf{Data}} & \multirow{2}{*}{\textbf{Method}} & \multicolumn{2}{c}{\textbf{NoC} $\downarrow$} & \multicolumn{3}{c}{\textbf{mIoU} $\uparrow$} & \multicolumn{2}{c}{\textbf{NSDS} $\uparrow$} \\
		\cmidrule(lr){3-4} \cmidrule(lr){5-7} \cmidrule(lr){8-9}
		& & \textbf{@85} & \textbf{@90} & \textbf{@1} & \textbf{@10} & \textbf{@20} & \textbf{@10} & \textbf{@20} \\
		\midrule
		\multirow{4}{*}{ssTEM} 
		& CDNet & 11.10 & 14.65 & - & 66.72 & - & - & - \\
		& RITM & 3.71 & 5.68 & - & 93.15 & - & - & - \\
		& SClick & 3.78 & 5.21 & 36.44 & \textbf{93.84} & 94.39 & 0.83 & 0.87 \\
		& Ours & \textbf{3.69} & \textbf{5.13} & \textbf{40.49} & 93.73 & \textbf{94.94} & \textbf{0.83} & \textbf{0.90} \\
		\midrule
		\multirow{4}{*}{BraTS} 
		& CDNet & 17.07 & 18.86 & - & 58.34 & 82.07 & - & - \\
		& RITM & \textbf{8.47} & \textbf{12.59} & - & \textbf{87.05} & 90.47 & - & - \\
		& SClick & 9.93 & 13.90 & 9.08 & 83.99 & 90.02 & 0.63 & 0.83 \\
		& Ours & 8.94 & 12.71 & \textbf{10.16} & 85.91 & \textbf{90.81} & \textbf{0.68} & \textbf{0.86} \\
		\midrule
		\multirow{4}{*}{OAIZIB} 
		& CDNet & 19.56 & 19.95 & - & 38.07 & 61.17 & - & - \\
		& RITM & 17.70 & 19.95 & - & \textbf{71.04} & 78.52 & - & - \\
		& SClick & 18.44 & 19.95 & 2.52 & 56.15 & 76.48 & 0.58 & 0.85 \\
		& Ours & \textbf{16.68} & \textbf{19.26} & \textbf{6.32} & 66.70 & \textbf{80.17} & \textbf{0.72} & \textbf{0.90} \\
		\midrule
		\multirow{4}{*}{\shortstack[l]{SEM-Precipitate}} 
		& CDNet & 16.26 & 18.64 & 27.09 & 60.85 & 77.98 & - & - \\
		& RITM & 9.57 & 14.48 & 19.93 & 82.82 & 85.38 & - & - \\
		& SClick & 10.37 & 14.92 & 10.73 & 82.16 & 86.26 & 0.81 & 0.89 \\
		& Ours & \textbf{7.92} & \textbf{12.21} & \textbf{27.51} & \textbf{85.74} & \textbf{89.46} & \textbf{0.87} & \textbf{0.93} \\
		\bottomrule
	\end{tabular}
	\caption{Out-of-domain generalization. All models were trained on natural images (SBD) and evaluated on datasets without fine-tuning. SClick: SimpleClick \cite{liu2023simpleclick},CDNet \cite{chen2021conditional}, RITM \cite{sofiiuk2021reviving}.}
	\label{tab:ood_compact}
\end{table}

\subsection{Quantitative Comparison with State-of-the-Art}
To validate the effectiveness of our proposed method, we conduct a comprehensive quantitative comparison against sixteen recent state-of-the-art (SOTA) approaches for interactive segmentation. They are trained on SBD dataset. As shown in \Cref{tab:comparision_sota_noc}, we evaluate the models on five standard benchmarks: GrabCut, Berkeley, SBD, DAVIS, and Pascal VOC. The primary metric for comparison is the NoC required to reach an IoU of 85\%, 90\%, and 95\%.

The base model, which incorporates the dual-head architecture, already demonstrates competitive performance, outperforming many existing methods. However, the most significant gains are realized with the introduction of our U-CFR framework. The final model, Ours (Base+U-CFR1), establishes a new state-of-the-art across nearly all evaluated metrics. Compared directly to the previous leading method, SimpleClick \cite{liu2023simpleclick}, our model shows remarkable improvements in user-click efficiency. On the Pascal VOC dataset, our model reduces the NoC@90 from 2.81 to 2.64, an improvement of 6.0\%. This efficiency gain is even more pronounced at the challenging NoC@95 threshold, where our model requires only 3.50 clicks compared to SimpleClick's 3.75, a 6.7\% reduction.

\begin{figure*}[!ht]
	\centering
	\includegraphics[width=\textwidth]{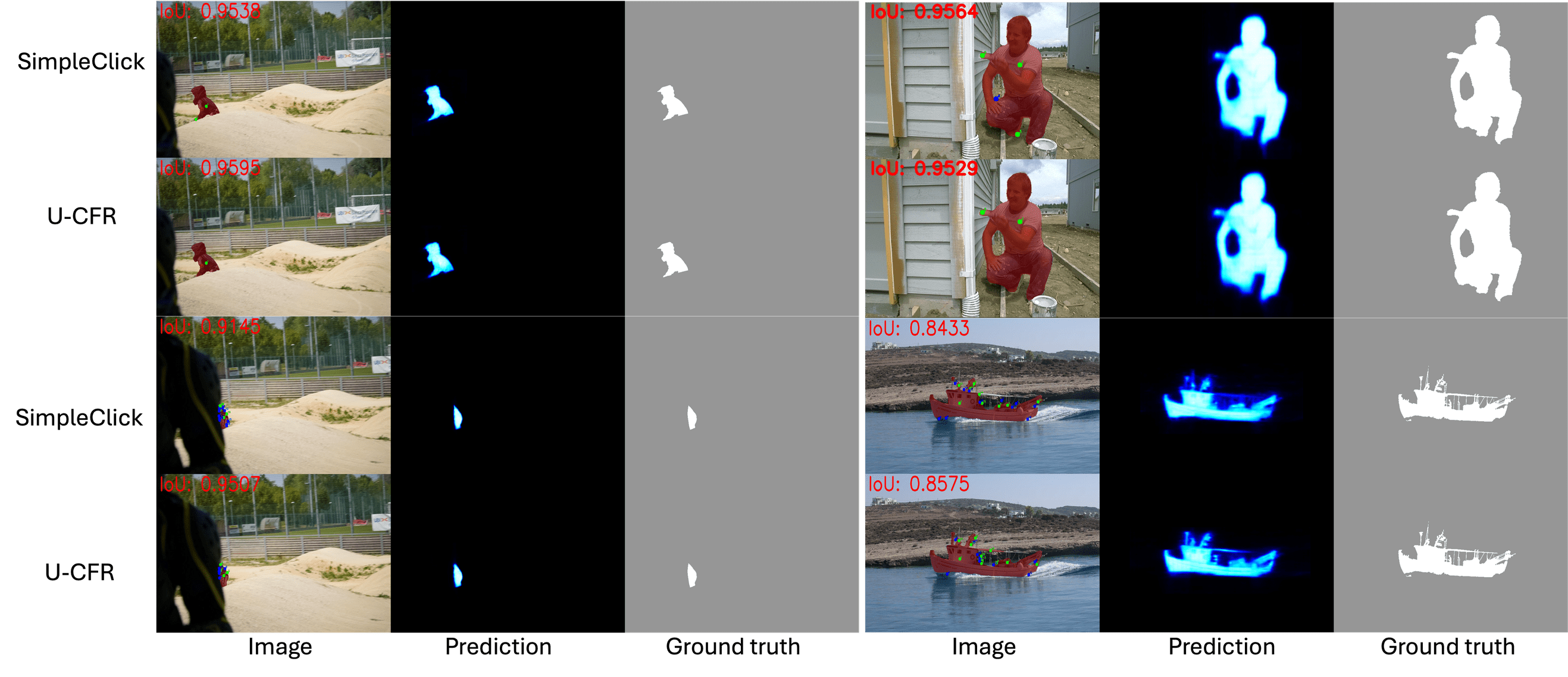}
	\caption{Qualitative comparison between our proposed U-CFR model and SimpleClick~\cite{liu2023simpleclick}.}
	\label{fig:segmentation}
\end{figure*}

This pattern is consistent with other benchmarks. In the Berkeley data set, we observe an 11\% improvement in NoC@90 (2.19 vs. 2.46) and a 9.5\% improvement in NoC@95 (6.07 vs. 6.71). Similarly, on SBD, our model is more efficient, reducing the NoC@90 by 5.7\% (4.94 vs. 5.24). Although SimpleClick \cite{liu2023simpleclick} achieves similar results on the GrabCut benchmark, our model is highly competitive and achieves the best performance at NoC@90 (1.50 clicks) and NoC@95 (1.82 clicks). \Cref{fig:segmentation} demonstrates results of U-CFR on four images, and U-CFR produces competitive segmentation results to the SimpleClick with fewer user clicks.

\begin{figure*}[!ht]
	\centering
	\includegraphics[width=0.7\linewidth]{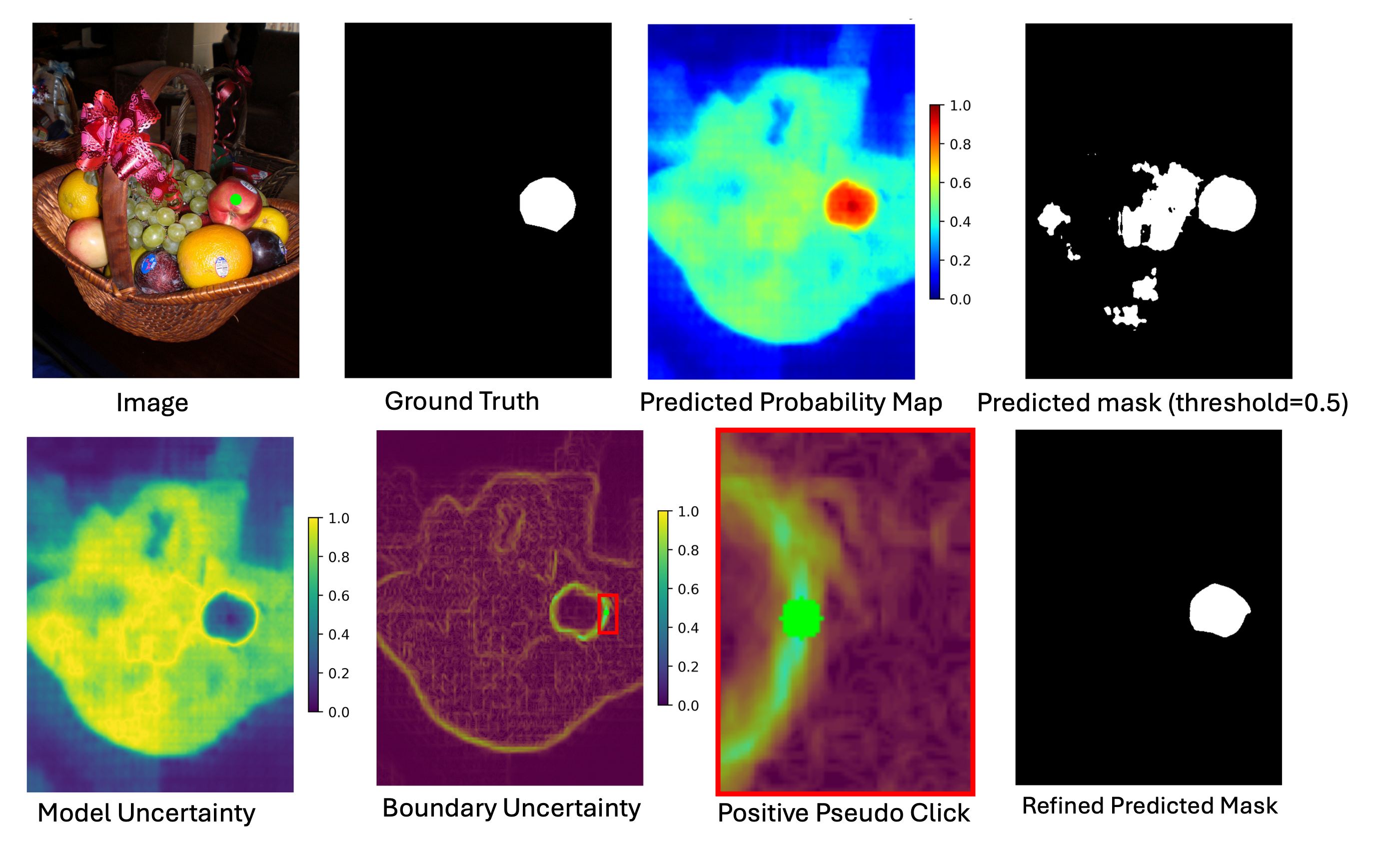}
	\caption{ Sample of U-CFR refinement. The input image (with a positive click) is processed to generate a prediction probability map and an initial binary mask.}
	\label{fig:ucfr}
\end{figure*}

Overall, the results in \Cref{tab:comparision_sota_noc}, \Cref{fig:segmentation}, \Cref{fig:ucfr} clearly indicate that our proposed framework, particularly with U-CFR, sets a new standard for interactive segmentation, requiring fewer clicks to achieve high-fidelity segmentation masks and thereby reducing the annotation burden on the user.


\subsection{Out-of-Domain Evaluation} 

To assess the generalization capability of our model, we extend our evaluation to datasets from two distinct and challenging domains. First, within the medical imaging domain, we utilize three established benchmarks: BraTS~\cite{baid2021rsna}, ssTEM~\cite{gerhard2013segmented}, and OAIZIB~\cite{ambellan2019automated}. Secondly, extending beyond biological imaging, we collaborate with the Idaho National Laboratory to evaluate performance on a unique, curated dataset of irradiated metallic SEM images for precipitate segmentation. Crucially, to test strict domain robustness, all models were trained solely on the SBD dataset and applied directly to these unseen medical and material science tasks without any fine-tuning. The results, which highlight our model's adaptability to significant domain shifts, are summarized in \Cref{tab:ood_compact}.
On the OAIZIB dataset, our model demonstrates a substantial leap in efficiency, reducing the NoC@85 from 18.44 (SimpleClick) to 16.68, an improvement of 9.5\%. For the initial mask quality, our model's mIoU@1 of 6.32\% is more than 150\% higher than SimpleClick's 2.52\%. This quality is further reflected in boundary accuracy, with our model achieving a remarkable 24\% relative improvement in NSDS at 10 clicks (0.72 vs. 0.58) and a higher final score at 20 clicks. These improvements are consistent across all the other medical datasets.
The SEM-Precipitate dataset presents a unique challenge due to the intricate and varying shapes of its objects. As shown in \Cref{tab:ood_compact}, our base model demonstrates superior performance compared to the baseline and competing methods. Specifically, our approach achieves a remarkable reduction in user effort, requiring only 7.92 clicks to reach 85\% IoU, compared to 10.37 for SimpleClick \cite{liu2023simpleclick} and 16.26 for CDNet \cite{chen2021conditional}. This efficiency is mirrored in the initial segmentation quality, where our model achieves a significantly higher mIoU of 27.51\% after just the first click, providing a much stronger starting point for refinement. Furthermore, our method consistently produces more accurate boundaries, as evidenced by the higher NSDS scores at both 10 clicks (0.87 vs. 0.81) and 20 clicks (0.93 vs. 0.89) compared to SimpleClick. These results highlight our model's robustness in handling complex boundary structures with high precision.

\section{Conclusion}

In this study, we introduce a dual-head architecture and an uncertainty-guided refinement strategy to advance the state-of-the-art in interactive image segmentation. The proposed approach makes two primary contributions: 1) the integration of an auxiliary EDH, which enriches the model's feature representations with explicit boundary awareness, and 2) the development of the U-CFR framework, an efficient inference-time strategy that uses uncertainty to automatically place pseudo-clicks and correct segmentation errors without additional user input. The experimental results demonstrated that the EDH-enhanced base model consistently outperforms strong baselines, showing marked improvements in click efficiency (NoC), initial mask quality (mIoU@1), and boundary accuracy (NSDS). This proves particularly beneficial in out-of-domain scenarios, where the model showed robust generalization on challenging medical datasets, generating significantly more accurate initial masks than prior work. Furthermore, the results confirmed that U-CFR provides an additional layer of refinement, measurably reducing the number of clicks required to achieve high-precision segmentation. By creating a model that is both more accurate from the first click and capable of self-correction, our work reduces the cognitive load and annotation effort required from human users. This research paves the way for more efficient, intuitive, and practical interactive segmentation tools. 


{\textbf{Acknowledgement} The authors acknowledge the financial support from the U.S. Department of Energy (DOE). The work was supported through the Idaho National Laboratory Laboratory Directed Research and Development (LDRD) Program under DOE Idaho Operations Office Contract DE-AC07-05ID14517, and through the Advanced Fuels Campaign (AFC) of the Nuclear Technology Research and Development program in the Office of Nuclear Energy under DE-AC07–05ID14517.}

%
%
%

\bibliographystyle{plainnat} 

\bibliography{references}

@String(PAMI = {IEEE Trans. Pattern Anal. Mach. Intell.})

@String(IJCV = {Int. J. Comput. Vis.})

@String(CVPR= {IEEE Conf. Comput. Vis. Pattern Recog.})

@String(ICCV= {Int. Conf. Comput. Vis.})

@String(ECCV= {Eur. Conf. Comput. Vis.})

@String(ICPR = {Int. Conf. Pattern Recog.})

@String(TOG= {ACM Trans. Graph.})

@String(TIP  = {IEEE Trans. Image Process.})

@String(ACMMM= {ACM Int. Conf. Multimedia})

@String(AAAI = {AAAI})

@String(PAMI  = {IEEE TPAMI})

@String(IJCV  = {IJCV})

@String(CVPR  = {CVPR})

@String(ICCV  = {ICCV})

@String(ECCV  = {ECCV})

@String(ICPR  = {ICPR})

@String(TOG   = {ACM TOG})

@String(TIP   = {IEEE TIP})

@String(TCSVT = {IEEE TCSVT})

@String(ACMMM = {ACM MM})

@inproceedings{hariharan2011semantic,
	title={Semantic contours from inverse detectors},
	author={Hariharan, Bharath and Arbel{\'a}ez, Pablo and others},
	booktitle={ICCV},
	@pages={991--998},
	year={2011},
	@organization={IEEE}
}

@inproceedings{lin2014microsoft,
	title={Microsoft COCO: Common objects in context},
	author={Lin, Tsung-Yi and Maire, Michael and others},
	booktitle={ECCV},
	pages={740--755},
	year={2014},
	@organization={Springer}
	}

@article{everingham2010pascal,
	title={The pascal visual object classes (voc) challenge},
	author={Everingham, Mark and Van Gool, Luc and Williams, Christopher KI and others},
	journal={IJCV},
	volume={88},
	@number={2},
	@pages={303--338},
	year={2010},
	@publisher={Springer}
	}

@article{rother2004grabcut,
	title={" GrabCut" interactive foreground extraction using iterated graph cuts},
	author={Rother, Carsten and Kolmogorov, Vladimir and Blake, Andrew},
	journal={TOG},
	@volume={23},
	@number={3},
	@pages={309--314},
	year={2004},
	@publisher={ACM New York, NY, USA}
	}

@inproceedings{martin2001database,
	title={A Database of Human Segmented Natural Image},
	author={Martin, David and Fowlkes, Charless and Tal, Doron and Malik, Jitendra},
	booktitle={ICCV},
	@volume={2},
	@pages={416--423},
	year={2001},
	@organization={IEEE}
	}

@inproceedings{perazzi2016benchmark,
	title={A benchmark dataset and evaluation methodology for video object segmentation},
	author={Perazzi, Federico and others},
	booktitle={CVPR},
	@pages={724--732},
	year={2016}
    }

@article{baid2021rsna,
	title={The RSNA-ASNR-MICCAI BraTS 2021 Benchmark},
	author={Baid, Ujjwal and others},
	journal={arXiv:2107.02314},
	year={2021}
	}

@article{gerhard2013segmented,
	title={Segmented anisotropic ssTEM dataset of neural tissue},
	author={Gerhard, Stephan and Funke, Jan and others},
	journal={figshare},
	@pages={0--0},
	year={2013},
	@publisher={figshare}
	}

@inproceedings{liu2023simpleclick,
	title={Simpleclick},
	author={Liu, Qin and Xu, Zhenlin and Bertasius, Gedas and Niethammer, Marc},
	booktitle={ICCV},
	@pages={22290--22300},
	year={2023}
	}

@article{dosovitskiy2020image,
	title={An image is worth 16x16 words: Transformers for image recognition at scale},
	author={Dosovitskiy, Alexey and others},
	journal={arXiv:2010.11929},
	year={2020}
	}

@INPROCEEDINGS{diceloss,
	author={Milletari, Fausto and Navab, Nassir and Ahmadi, Seyed-Ahmad},
	booktitle={3DV}, 
	title={V-Net: Fully Convolutional Neural Networks for Volumetric Segmentation}, 
	year={2016},
	@volume={},
	@number={},
	@pages={565-571},
	@keywords={Image segmentation;Feature extraction;Biomedical imaging;Three-dimensional displays;Neural networks;Magnetic resonance imaging;Two dimensional displays;Deep learning;segmentation;prostate;machine learning;convolutional neural networks},
	@doi={10.1109/3DV.2016.79}}

@inproceedings{kirillov2023segment,
	title={Segment anything},
	author={Kirillov, Alexander and others},
	booktitle={ICCV},
	@pages={4015--4026},
	year={2023}
	}

@inproceedings{chen2022focalclick,
	title={Focalclick},
	author={Chen, Xi and Zhao, Zhiyan and Zhang and others},
	booktitle={IEEE/CVF CVPR},
	@pages={1300--1309},
	year={2022}
	}

@inproceedings{sun2024cfr,
	title={CFR-ICL},
	author={Sun, Shoukun and Xian, Min and others},
	booktitle={AAAI},
	@volume={38},
	@number={5},
	@pages={5017--5024},
	year={2024}
	}

@article{zou2023segment,
	title={Segment everything everywhere all at once},
	author={Zou, Xueyan and others},
	journal={NeurIPS},
	@volume={36},
	@pages={19769--19782},
	year={2023}
	}

@inproceedings{liew2021deep,
	title={Deep interactive thin object selection},
	author={Liew, Jun Hao and others},
	booktitle={WACV},
	@pages={305--314},
	year={2021}
	}

@article{sang2022small,
	title={Small-object sensitive segmentation using across feature map attention},
	author={Sang, Shengtian and others},
	journal={PAMI},
	@volume={45},
	@number={5},
	@pages={6289--6306},
	year={2022},
	@publisher={IEEE}
	}

@inproceedings{he2016deep,
	title={Deep Residual Learning},
	author={He, Kaiming and others},
	booktitle={CVPR},
	@pages={770--778},
	year={2016}
	}

@article{wang2020deep,
	title={Deep High-Resolution Representation Learning},
	author={Wang, Jingdong and others},
	journal={PAMI},
	@volume={43},
	@number={10},
	@pages={3349--3364},
	year={2020},
	publisher={IEEE}
	}

@inproceedings{he2022masked,
	title={Masked autoencoders are scalable vision learners},
	author={He, Kaiming and others},
	booktitle={CVPR},
	@pages={16000--16009},
	year={2022}
	}

@article{wang2024order,
	title={Order-Aware Interactive Segmentation},
	author={Wang, Bin and others},
	journal={arXiv:2410.12214},
	year={2024}
	}

@article{xu2025mst,
	title={MST: Adaptive Multi-Scale Tokens Guided Interactive Segmentation},
	author={Xu, Long and Chen, Yongquan and Li, Shanghong and others},
	journal={TCSVT},
	year={2025},
	@publisher={IEEE}
	}

@inproceedings{khattar2021cross,
	title={Cross-Domain Multi-Task Learning },
	author={Khattar, Apoorv and Hegde, Srinidhi and Hebbalaguppe, Ramya},
	booktitle={CVPR},
	@pages={3639--3648},
	year={2021}
	}

@inproceedings{wang2021boundary,
	title={Boundary-Aware Transformers},
	author={Wang, Jiacheng and others},
	booktitle={MICCAI},
	pages={206--216},
	year={2021},
	@organization={Springer}
	}

@article{wang2023segrefiner,
	title={SegRefiner: Towards model-agnostic segmentation refinement with discrete diffusion process},
	author={Wang, Mengyu and others},
	journal={arXiv:2312.12425},
	year={2023}
	}

@inproceedings{liu2022interactive,
	title={Interactive image segmentation with click imitation},
	author={Liu, Qin and others},
	booktitle={ECCV},
	@pages={23--27},
	year={2022}
	}

@article{paszke2019pytorch,
	title={Pytorch},
	author={Paszke, Adam and others},
	journal={NeurIPS},
	volume={32},
	year={2019}
	}

@inproceedings{lin2022focuscut,
	title={FocusCut},
	author={Lin, Zheng and Duan, Zheng-Peng and others},
	booktitle={CVPR},
	pages={2637--2646},
	year={2022}
	}

@inproceedings{li2018interactive,
	title={Interactive image segmentation with latent diversity},
	author={Li, Zhuwen and Chen, Qifeng and others},
	booktitle={CVPR},
	pages={577--585},
	year={2018}
	}

@inproceedings{jang2019interactive,
	title={Interactive image segmentation via backpropagating refinement scheme},
	author={Jang, Won-Dong and Kim, Chang-Su},
	booktitle={CVPR},
	pages={5297--5306},
	year={2019}
	}

@inproceedings{sofiiuk2020f,
	title={F-BRS},
	author={Sofiiuk, Konstantin and Petrov, Ilia and Barinova, Olga and others},
	booktitle={CVPR},
	pages={8623--8632},
	year={2020}
	}

@article{sofiiuk2021reviving,
	title={Reviving iterative training with mask guidance for interactive segmentation},
	author={Sofiiuk, Konstantin and Petrov, Ilia A and Konushin, Anton},
	journal={arXiv:2102.06583},
	year={2021}
	}

@inproceedings{chen2021conditional,
	title={Conditional diffusion for interactive segmentation},
	author={Chen, Xi and Zhao, Zhiyan and Yu, Feiwu and others},
	booktitle={ICCV},
	@pages={7345--7354},
	year={2021}
	}

@inproceedings{xian2016eiseg,
	title={EISeg: Effective interactive segmentation},
	author={Xian, Min and others},
	booktitle={ICPR},
	@pages={1982--1987},
	year={2016},
	organization={IEEE}
	}

@article{xian2016neutro,
	title={Neutro-connectedness cut},
	author={Xian, Min and others},
	journal={TIP},
	@volume={25},
	@number={10},
	@pages={4691--4703},
	year={2016},
	publisher={IEEE}
	}

@inproceedings{zhang2020interactive,
	title={Interactive object segmentation with inside-outside guidance},
	author={Zhang, Shiyin and others},
	booktitle={CVPR},
	@pages={12234--12244},
	year={2020}
	}

@article{xu2017deep,
	title={Deep GrabCut},
	author={Xu, Ning and others},
	journal={arXiv:1707.00243},
	year={2017}
	}

@article{kanopoulos1988design,
	title={Design of an image edge detection filter using the Sobel operator},
	author={Kanopoulos, Nick and Vasanthavada, Nagesh and others},
	journal={IEEE JSSC},
	volume={23},
	@number={2},
	@pages={358--367},
	year={1988},
	@publisher={IEEE}
	}

@inproceedings{sudre2017generalised,
	title={Generalised Dice Overlap},
	author={Sudre, Carole H and others},
	booktitle={DLMIA},
	pages={240--248},
	year={2017},
	organization={Springer}
	}

@article{kavur2021chaos,
	title={CHAOS challenge-combined (CT-MR) healthy abdominal organ segmentation},
	author={Kavur, A Emre and others},
	journal={MedIA},
	@volume={69},
	pages={101950},
	year={2021},
	@publisher={Elsevier}
	}

@inproceedings{zhou2023interactive,
	title={Interactive segmentation as gaussion process classification},
	author={Zhou, Minghao and others},
	booktitle={CVPR},
	@pages={19488--19497},
	year={2023}
	}

@inproceedings{wei2023focused,
	title={Focused and collaborative feedback integration for interactive image segmentation},
	author={Wei, Qiaoqiao and others},
	booktitle={CVPR},
	@pages={18643--18652},
	year={2023}
	}

@inproceedings{du2023efficient,
	title={Efficient mask correction for click-based interactive image segmentation},
	author={Du, Fei and Yuan, Jianlong and others},
	booktitle={CVPR},
	@pages={22773--22782},
	year={2023}
	}

@inproceedings{zeng2023feature,
	title={Feature decoupling-recycling network for fast interactive segmentation},
	author={Zeng, Huimin and others},
	booktitle={ACMMM},
	pages={6665--6675},
	year={2023}
	}

@article{ambellan2019automated,
	title={Automated Segmentation of knee bone and cartilage combining statistical shape and CNN},
	author={Ambellan, Felix and Tack, Alexander and others},
	journal={MedIA},
	@volume={52},
	@pages={109--118},
	year={2019},
	@publisher={Elsevier}
	}

@inproceedings{deng2018learning,
	title={Learning to predict crisp boundaries},
	author={Deng, Ruoxi and Shen, Chunhua and Liu, Shengjun and others},
	booktitle={ECCV},
	@pages={562--578},
	year={2018}
	}

@inproceedings{sofiiuk2019adaptis,
	title={Adaptis: Adaptive instance selection network},
	author={Sofiiuk, Konstantin and Barinova, Olga and Konushin, Anton},
	booktitle={ICCV},
	@pages={7355--7363},
	year={2019}
	}

@inproceedings{lin2017feature,
	title={Feature pyramid networks for object detection},
	author={Lin, Tsung-Yi and others},
	booktitle={CVPR},
	@pages={2117--2125},
	year={2017}
	}

@inproceedings{shit2021cldice,
	title={clDice-a novel topology-preserving loss function for tubular structure segmentation},
	author={Shit, Suprosanna and Paetzold, Johannes C and Sekuboyina, Anjany and others},
	booktitle={Proceedings of the IEEE/CVF conference on computer vision and pattern recognition},
	pages={16560--16569},
	year={2021}
	}
\end{document}